# A Survey on VQA: Datasets and Approaches

Yeyun Zou[†]
School of Advanced Technology
Xi'an Jiaotong-Liverpool University
SuZhou, JiangSu
Yeyun.Zou17@student.xjtlu.edu.cn

Qiyu Xie[†]
College of Liberal Arts and Sciences
University of Iowa
Iowa City, Iowa
[*]qiyu-xie@uiowa.edu
[†]These authors contributed equally.

*Abstract*—Visual question answering (VQA) is a task that combines both the techniques of computer vision and natural language processing. It requires models to answer a text-based question according to the information contained in a visual. In recent years, the research field of VQA has been expanded. Research that focuses on the VQA, examining the reasoning ability and VQA on scientific diagrams, has also been explored more. Meanwhile, more multimodal feature fusion mechanisms have been proposed. This paper will review and analyze existing datasets, metrics, and models proposed for the VQA task.

## I INTRODUCTION

Visual question answering (VQA) task was initially proposed in S. Antol et al.'s research [1]. It requires the model to take both the question Q in natural language and the image I as input and generate the answer A according to the information contained in the inputs. The subtasks of VQA could be divided into two categories: one could be completed with ground truth information in the image, and the other requires reasoning according to the knowledge beyond the image. The former one includes fine-grained recognition (e.g., "What kind of cheese is on the pizza?"), scene recognition (e.g., "What is the weather today?"), activity recognition (e.g., "Is the girl walking?"), object detection (e.g., "Is there a tree in the image?"), attribute classification (e.g., "What color are her eyes?") and counting (e.g., "How many birds are in the image?"). The latter one consists of knowledge-based reasoning (e.g., "Is it a vegetarian sandwich?"), commonsense reasoning (e.g., "Why she cries?") and spatial relationship recognition (e.g., "Is the dog in front of the house?").

In recent years, visual question answering has attracted the attention of researchers from computer vision, natural language processing, knowledge representation, and other machine learning communities. As a result of the flourish in this field, datasets, metrics, and models have been proposed, and the scope of research has been expanded. For instance, the visual contents provided in the datasets have been enriched by videos when many video question answering datasets, such as TVQA [2], MovieQA[3], were proposed. Meanwhile, as visual question answering is deemed as a proxy for AI-complete problems, datasets aiming to examine the knowledge-based reasoning ability of models have been promoted. Moreover, the knowledge-based video question answering datasets were also created to test the reasoning ability of models in complex and continuous scenes. Besides, to overcome the statistical bias and language prior in the existing VQA datasets, datasets like VQA v2[4] and VQA-CP [5] were established. VQA could also be applied in scientific diagrams analysis, while datasets consisting of various scientific figures, such as DVQA [6] and Figure-QA [7], were created.

Models have been constructed to accomplish the tasks proposed by the datasets and solve the problems emerging in previous works. To overcome the language prior, models, like the decomposed linguistic representation, have been constructed. Aiming to improve accuracy, many state-of-the-art mechanisms were introduced and combined with current VQA techniques. For instance, according to the BERT [8], pre-train models such as LXMERT [9] and ViLBERT[10] were constructed, and adversarial learning was also applied in IFGSM[11] and the paraphrasing[12] model to improve the accuracy. In addition, to enhance the efficiency of visual question answering, the multimodal information fusion mechanisms such as BLOCK [13], grid-feature[14], and DACT[15] were proposed. Besides, based on the video question answering datasets and scientific diagram based datasets, video question answering models, such as ROLL [16] and hstar[17], as well as scientific diagram analyzing models have been established.

This paper would review the existing datasets, metrics, and models of VQA and analyze their progress and remaining problems.

## II DATASET

As VQA task has attached more attention to researchers in recent years, the diversity of datasets for visual question answering has increased. For instance, VQA v2[18] and VQA-CP [19] were proposed to eliminate the language prior and enhance the visual understanding ability of the model. Compared to datasets contains comprehensive contents such as VQA [20], Visual Genome [21], Flickr30k [22], etc., some datasets that concentrate on the specific scenario were also published. For instance, FigureQA [7] is an image dataset that consists of scientific diagrams, aiming to propel the research on the visual understanding of the statistical figure. Social-IQ [23] is a video dataset aimed to enhance the sentiment detection of the model. Besides, knowledge-based datasets have been developed recently because knowledge-based reasoning tasks and commonsense reasoning tasks have attached more attention. Hence, R-VQA [24], FVQA [25], KVQA [26], etc. have been proposed. When image datasets have been widely explored, video datasets were also published, such as MovieQA[3], PororoQA[27], TVQA[28], etc. In the following subsections, these datasets would be introduced and analyzed.

## A. Datasets of Images

### 1) DAQUAR[29]

The Dataset for Question answering on real-world images was built based on the NYU-Depth V2 dataset [30]. The data were annotated through two methods: synthetic and human. The synthetic question-answer pairs were generated automatically based on templates, while the human question-answer pairs were collected by in-house participants. For the question-answer pairs generated by the human, biases exist because the attention mechanism leads people to focus on the prominent objects of the image.

### 2) Flickr30k entities[22]

After the Flickr30k dataset becoming a widely accepted benchmark for visual question answering tasks, Flickr30k entities were augmented to facilitate the training and tests. Flickr30k entities contain 31,783 images that mainly focus on humans and animals, and each image owns five English captions on average. It identifies which captions of the same images refer to the same set of entities. Hence, 244,035 conference chains and 275,775 bounding boxes that extract the entities in the image were generated.

### 3) VQA[20]

VQA is a comprehensive dataset that contains 204,721 images from MS COCO [31] and 50,000 scenes from an abstract dataset [32]. Both open-ended and multiple-choice question-answer pairs are collected for visual question answering tasks. Several question-answer pairs were assigned to one image in this dataset. Commonsense knowledge is required to answer any questions in the VQA dataset, while many questions only ask for ground-truth answers. It has already become a standard benchmark for the VQA task.

### 4) Visual Genome[21]

Visual Genome aims to enhance the progress on cognitive tasks, especially spatial relationship reasoning. The dataset contains over 108,000 images, which have an average of 35 objects, 26 attributes, and 21 pairwise relationships between objects. It attaches the importance of relationships and attributes in annotation space because they are essential elements for visual understanding. Hence, it collects more than 50 descriptions for different components of the image in this dataset.

### 5) Visual7W[33]

Visual7W applies the six W questions (what, where, when, who, why, and how) to systematically examine the model's capability for visual understanding and append "which" question category. Questions in the dataset were standardized into the multiple-choice format. There are four candidates for each question, and only one candidate is the correct answer.

### 6) Visual Madlibs[34]

Visual Madlibs is a dataset consisting of 360,001 targeted descriptions spanned from 12 different types of templates and their corresponding images. With this dataset, more fine-grained and specific descriptions could be generated so that the model could be asked more detailed questions. Moreover, questions about aspects beyond the scope of ground-truth information are depicted in the image.

### 7) FM-IQA[35]

FM-IQA is a large-scale multilingual visual question answering dataset. It was generated from MS COCO[31]. There are 158,392 images with 316,193 Chinese question-answer pairs and their English translations. Each image has at least two question-answer pairs as annotations. The average lengths of the questions and answers are 7.38 and 3.82, respectively. 1,000 question-answer pairs and their corresponding image are randomly added to the test set.

## B. Balanced VQA

The statistical bias and the language priors existing in the dataset would interfere with the performance evaluation because the model could cheat to get better results by exploiting the language structure of questions and corresponding statistical patterns of the answers. For instance, in VQA[20] dataset, "2" is the correct answer for 39% of the questions start with "How many"[18]. Through this trick skill, the model trained on the datasets with language prior and statistical bias could answer questions without understanding the image. To solve this problem, some datasets such as VQA2.0[18], were proposed.

### 1) Binary Visual Question Answering on the Abstract Scene[36]

This dataset eliminates the statistical bias and language priors in the binary visual question answering tasks by creating pairs of complementary scenes and converting questions into a tuple that summarizes the relationship between two objects. Hence, the task of answering binary questions is translated into visual verification because the model only needs to detect whether an object exists in the image and return "yes" or "no". The performance of some state of art models is worse on this dataset than VQA. This phenomenon proves that language prior causes the performances of some state-of-art models are overrated.

### 2) VQA v2.0[18]

VQA v2.0 is established on the hypothesis that a balanced dataset would force the model to focus on visual information. It balanced the VQA [20] dataset by giving an (image, question, answer) triplet (I, Q, A) from the VQA dataset and asking humans to identify an image I' that is similar to I but results in answers to the question Q become A'. Hence, the model would be forced to understand the information contained in the image because the same question has two different answers for two different images. VQA v2.0 consists of 1.1 million (image, question) pairs with 13 million associated answers. Compared with binary VQA dataset[36], VQA v2.0 is more comprehensive because it generated balanced real-world images from MS COCO [31] rather than only focus on abstract scenes, and it extends the questions from binary questions to all questions.

*3) VQA-CP v1 & VQA-CP v2[5]*

The VQA-CP v1 and VQA CP-CP v2 splits are created to reduce the influence of statistical bias existing in answers, by differing the distribution of answers for each question type in the training set and test set. They split and re-organized the data in the VQA v1 dataset and VQA v2 dataset, respectively, by question grouping and greedily re-splitting methods. In question grouping, questions generated for the same task and corresponding to the same ground-truth answer are grouped together. For instance, ("What color is the dog", white) and ("what color is the bird", while) were grouped together while ("What color is the dog", white) and ("What color is the dog", yellow) were in the different group. Subsequently, a greedy approach was designed to redistribute the data into train and test sets, considering maximizing the number of concepts and preventing groups from repeating between train and test set. The train set of VQA-CP v1 consists of 118K images, 245K questions, and 2.5 Million answers, and the test set comprises 87K images, 125K questions and 1.3 Million answers. In VQA-CP v2, there are 121K images, 438K questions and 4.4 million answers for trains and 98K images, 220K questions, and 2.2 million answers for tests. The degraded performance of baseline models on VQA-CP demonstrates that this dataset provided a relatively balanced environment for VQA models.

*C. Datasets of Statistical Figures*

In recent years, many datasets that concentrate on scientific figures emerged and propel the research on the understanding of statistical properties contained in the diagrams. The visual question answering tasks for diagrams differ from the general visual question answering task in many aspects. One difference is that the attributes of objects have different importance for natural images and scientific figures, especially for the color and the area. Both the color and the area only attribute that has an impact on the questions about the properties of objects in the natural image while they have a unique meaning in the scientific diagrams. For instance, the same color in a bar chart refers to the objects in the same category, and the area of the bar matches the number of objects under the same category. Hence, the published datasets of scientific diagrams provide a chance to develop models regarding the uniqueness of the statistical figures.

*1) DVQA[6]*

DVQA is a dataset that focuses on bar chart understanding. It consists of charts from scientific documents, webpages, and business reports and over 3 million image-question pairs. The dataset is designed for testing three diagram understanding subtask: structure understanding, data retrieval, and reasoning. It tries to mitigate the statistical bias by randomly removing questions that have strong priors until the number of yes/no answers towards each question is balanced.

*2) FigureQA[7]*

FigureQA is a synthetic figure dataset including over 100,000 lines, dot-lines, vertical bars, horizontal bars, and pie plots. Questions concern one-to-all and one-to-one resolution require inference among plot elements. The dataset consists of 15 different types of questions around the statistical properties, such as maximum, minimum, median, smoothness, the relationship between two numbers, and the intersection. Each question corresponds to an answer, which is either yes or no.

*3) LEAF-QA[37]*

LEAF-QA is a comprehensive visual question answering corpus that includes 250,000 densely annotated figures and charts collected from a real-world open data source, such as government census and financial data, along with 1.5 million questions and 2 million answers. LEAF-QA consists of various plots, including the group bar, snack bar, pie, donut, box, line, and scatter. The questions in this dataset are generated based on analytical reasoning questions in the Graduate Record Examination. The answers could be divided into three categories: chart vocabulary (i.e., answers contained in the chart such as seeking the label with the maximum value in the chart), common vocabulary (i.e., answers include common words such as yes or no and answers towards counting problem) and chart type (i.e., the answers are type name of charts).

*D. Knowledge-based Datasets*

As a visual question answering task is deemed as a proxy of evaluating the understanding ability of a system, the existing effort on VQA that focuses on generate ground-truth answers is inadequate for the "AI-complete" task. For instance, a conventional visual question answering system could identify that the red object in the image is a fire hydrant but unaware that the fire hydrant is used to stop the fire from spreading. Comapred to conventional VQA tasks, knowledge-based VQA task is more challenging. It requires the model to identify the necessary knowledge, find it in the knowledge base and incorporate the knowledge, the image feature and the question representation to answer the question. Several knowledge-based datasets have been proposed to promote the development of knowledge-based reasoning tasks and provide a benchmark for evaluation.

*1) KBVQA*

The goal of the KB-VQA dataset is to construct a benchmark for evaluating the performance of VQA models on higher knowledge-level questions and explicit reasoning tasks with provided external knowledge. To build the dataset, 700 images in MS COCO [31] covering around 150 object classes and 100 scene classes were selected. For images, 2,402 question-answering pairs were generated by human questioners ranging from object identification questions, attribute detection questions to commonsense reasoning questions (i.e., do not require to refer to an external source) and knowledge-based reasoning questions (such as "When was the home appliance in this image invented?").

*2) FVQA[25]*

Besides images and question-answer pairs, the Fact-based VQA dataset additionally provides support-fact, which is a structured representation of knowledge stored in external KBs and indispensable for answering a given visual question. This dataset sampled 2,190 images from MS COCO [31] and extracted three types of objects: Object (i.e., named entities such as people, dog, and tree), Scene (such as office, bedroom, and beach), and Action (such as swimming, jumping and surfing). The knowledge was extracted from DBpedia[38],

ConceptNet[39] and WebChild[40]. In total, 5,826 questions corresponding to 4,216 unique facts are collected.

*3) KVQA[26]*

Knowledge-aware VQA dataset contains 183,007 question-answer pairs of about 18,000 people within 24,602 images. Multi-entity, multi-relation, and multi-hop reasoning are required to answer the questions in the KVQA dataset. Compared with KB-VQA and FVQA, KVQA not only owns a larger size but leads to the problem of visual entity linking where the task is to link the named entity, appearing in an image to one of the entities in Wikidata. It enables the visual named entity linking by providing a support set containing reference images of 69,000 persons from Wikidata.

*4) R-VQA[24]*

Relation-VQA is built on Visual Genome [21] dataset. It includes 335,000 data samples, and each sample consists of an (image, question, answer) pair and an aligned supporting relation fact. The relation facts comprise three different types: entity concept (there, is, object), entity attribute (subject, is, attribute), and entities' relation (subject, relation, object) based on the annotated semantic data of concepts, attributes, and relationships in visual Genome.

*5) OK-VQA[41]*

Outside Knowledge, the VQA dataset comprises 12,951 unique questions out of 14,055 total, and 14,031 images selected from MS COCO[ 31]. The knowledge contained in this dataset is commonsense knowledge, including ten categories: Vehicles and Transportation; Brands; Companies and Products; Objects; Materials and Clothing; Sports and Recreation; Cooking and Food; Geography; History; Language and Culture; People and Everyday Life; Plants and Animals; Science and Technology and Weather and Climates.

*E. Datasets of Videos*

In recent years, researchers have tried to expand the visual question answering task from discrete images to continuous videos. Compared with images, the video question answering task requires a higher level of understanding ability and multimodal information fusion ability, especially in complex scenes. To enhance the development of the video VQA task, several datasets were proposed, ranging from movies, dramas social scenes that are highly associated with the real-life environment to abstract cartoons.

*1) MovieQA[42]*

MovieQA is a large-scale question answering dataset, aiming to create a benchmark for evaluating semantic understanding over long temporal data. It comprises 14,944 multiple-choice questions with five candidate answers where only one is correct. The questions range from the ones that only require direct observation to the "Why" and "How" questions that require reasoning. Timestamp annotations are also included in the dataset to help locate the questions and answers in the video. This dataset also includes plot synopses, video subtitles, DVS, and scripts as guidelines.

*2) TVQA[2]*

TVQA dataset was built on 6 popular TV shows that could be divided into 3 genres: medical dramas (such as House), sitcoms (such as The Big Bang Theory), and crime shows (such as Castle) with 152,5,000 human-written QA pairs. It contains 21,793 video clips from 925 episodes spanning 461 hours. Each video clip is 60-90 seconds long and associates with 7 questions and 5 candidates' answers (only one is correct). Questions in this dataset are structured in the format of [What/How/Where/Why/…] __ [when/before/after] ______ and require free-form answers.

*3) TVQA+[28]*

TVQA+ is a large-scale spatio-temporally grounded video question answering dataset augmented on the TVQA dataset. It provides temporal annotations that denote which parts of the video clip is required to answer the question by associating the bounding boxes with corresponding objects. This dataset comprises of 29,400 multiple-choice questions in both temporal and spatial domains and 310,800 bounding boxes linked with objects in 2,500 categories.

*4) PororoQA[27]*

Compared with movies and dramas, cartoon videos own a simpler story structure and a smaller environment as well as more abstract scenes and characters. PororoQA dataset was generated on a cartoon series comprised of 171 episodes whose average length is 7.2 minutes. This series was designed for kids so that it contains only ten main characters and the size of its vocabulary is around 4,000. The dataset includes 16,066 scene-dialogue pairs, 27,328 descriptive sentences, and 8,913 QA pairs. The questions in this dataset could be subdivided into 11 categories: action, person, abstract, detail, method, reason, location, statement, causality, yes/no, and time. For each question, a scene-dialogue pair would be attached as the knowledge base to help generate the answer.

*5) Social-IQ[23]*

The Social-IQ dataset aims to propel the research on real-life social situation understanding and explore the psychometric measurement on social intelligence. It was built on YouTube videos that cover various social and behavioral situations, such as birthday parties or basketball game. This dataset consists of 1,250 videos, 7,500 questions, and 52,500 candidates' answers, which comprise 30,000 correct answers and 22,500 incorrect answers. 50% of questions in the dataset are "Why" questions and "How" questions, which require reasoning ability. Because the Social-IQ dataset focuses on analyzing the social interactions, it also contains many "who" and "what" questions. For each question, 4 correct answers and 3 incorrect answers are provided. Hence, multiple explanations for a specified question in a social situation is allowed.

*6) KnowIT VQA[43]*

KnowIT-VQA is a combination of a knowledge-based visual question answering task and the video question answering task. It contains 12,087 clips split from the first nine seasons of the Big Bang Theory TV show with subtitles annotated to its timestamp and speaker. Besides the clips whose length is 20 seconds, this dataset also includes 24,282 question-answer pairs and specific or recurrent knowledge. The questions in this dataset are multi-choice questions and could be categorized into four types: visual-based (i.e., its answer could be found in video clips), textual-based (i.e., its answer could be found in subtitles), temporal-based (i.e., its answer could be obtained from the current video at a specific time), and knowledge-based (i.e., its answer is not contained in the current video but could be found in another sequence of the show). The specific knowledge and the recurrent knowledge are distinguished by whether the knowledge is from the same episode or it repeatedly appears during the show.

*F. Reasoning Ability Diagnostics Dataset*

The reasoning ability of the vision system was firstly systematically analyzed by CLEVR [44] dataset, which contains 100,000 images, and 999,968 questions. In the CLEVR, images are simple three-dimension figures, and information contained in each image is elusive and complete. These features of the dataset promote the models that own strong reasoning abilities to be proposed. Subsequently, RAVEN [45] was proposed in 2019 to propel the reasoning ability to evolve to a higher level. It consists of 1,120,000 images and 70,000 RPM problems that are widely accepted to be highly correlated with real intelligence. In addition, the problems were labeled in tree structures, and a total of 1,120,000 labels were contained in the dataset. Besides, 5 rule-governing attributes and 2 noise attributes were designed to attack visual systems' major weaknesses in short-term memory and compositional reasoning.

## III METRICS

*A. Simple Accuracy*

The questions in the visual question answering dataset could be divided into two categories: open-ended questions that require an answer in natural language and multiple-choice questions that need the model to select one correct answer from candidate answers. For the multiple-choice questions, it is efficient to utilize simple accuracy by calculating the ratio of right answers and total answers to evaluate the performance of the model. Simple accuracy could also be applied to evaluate the open-ended questions, but it could cause some problems. This is because the punishment could not reflect the magnitude of semantic differences. For instance, the answer to "What is in the sea?" might be salmon. If a model generated an answer that is "fish", it would be penalized as same as the answer "No". Hence, some improvements on metrics were proposed to improve the precision of metrics.

*B. VQA Accuracy[20]*

VQA accuracy was proposed in S. Antol et al.'s research [20] to evaluate the answers generated for the open-ended task. An answer is deemed 100% accuracy if at least 3 people provided the exact same answer. It is defined as follow:

$$accuracy = \min\left(\frac{n}{3}, 1\right) \quad (1)$$

There are several problems within the metrics. Firstly, the human annotators could not gain the agreements on answers and it restricts the highest accuracy that a model could obtain. For the "Why" questions, over 59% did not receive the same answers from more than 3 human annotators. In addition, an answer would get 1/3 accuracy in the worst situation. Furthermore, the VQA accuracy could lead to some mistakes on the "yes/no" questions. The answer, which is "yes" or "no", could repeat more than 3 times for a question. Hence, either the answer "yes" or the answer "no" could be well evaluated.

*C. Modified WUPS Score [29]*

Modified WUPS Score was proposed to measure the semantic difference between answers generated by the model and label provided by the human annotators. It was established on the Fuzzy Sets[46] and the WUP score[47]. The WUP score evaluates the similarity of two words by traversing the semantic tree, counting the number of nodes on the path between two words, and calculating the common subsumer. After that, a score between 0 and 1 would be assigned to the word pair. For instance, the WUP score of (curtain, blinds) pairs is 0.94. The larger the score is, the similar the two words are. Under this metric, the answer, which is more semantically similar to the label, would be penalized less.

The WUPS Score is defined as follows:

$$WUPS(A,T) = \frac{1}{N}\sum_{i=1}^{N} \min\left\{\prod_{a \in A^i} \max_{t \in T^i} WUP(a,t), \prod_{t \in T^i} \max_{a \in A^i} WUP(a,t)\right\} \cdot 100 \quad (2)$$

However, under this metric, a relatively high score would be assigned to distant words. Hence, the threshold was set to decide when to decrease the weight of WUP score. If the WUP score is less than the threshold, the weight would be downsized by multiplying a factor (0.1 is recommended.). The modified WUPS scores were applied to evaluate the model on the DAQUAR datasets. However, it also has some weaknesses. Firstly, it could give high scores to the answers that are semantically similar to the label but have different meanings. The problem has a negative influence on the attribute detection tasks. For example, red and pink are semantically similar, but for a question that requires the model to identify the color of a red fire hydrant, the answer pink should not receive a high score. In addition, the WUPS score could only be used to evaluate the similarities between discrete concepts. It cannot be applied to the questions whose answers are phrases or sentences such as the "Why" questions.

*D. Consensus [48]*

This metric was proposed to assign the answer that preferred by more human annotators higher priority. Two consensuses were proposed for the DAQUAR[29] dataset. One is the average consensus[48] and the other is min consensus[48]. The average consensus is defined as:

$$\frac{1}{NK}\sum_{i=1}^{N}\sum_{k=1}^{K} \min\left\{\prod_{a \in A^i} \max_{t \in T_k^i} \mu(a,t), \prod_{t \in T_k^i} \max_{a \in A^i} \mu(a,t)\right\} \quad (3)$$

Where $A^i$ is the answer towards the i-th question and $T_k^i$ is the k-th answer provided by the human corresponding to the k-th question.

And the min consensus is defined as:

$$\frac{1}{N}\sum_{i=1}^{N} \max_{k=1\ to\ K}\left(\min\left\{\prod_{a \in A^i}\max_{t \in T_k^i}\mu(a,t), \prod_{t \in T_k^i}\max_{a \in A^i}\mu(a,t)\right\}\right) \quad (4)$$

Compared with the average consensus, which ranks the consensus by the popularity among human annotators, the min consensus, the min consensus only needs the answer to have the consensus with one human annotator.

*E. Manual Test*

Manual test means that all the answers are evaluated by the human workers. For instance, the models on the dataset FM-IQA[35] were evaluated by the human and each person provided scores ranged from 0 to 2 for each answer to rank their degree of correctness. Compared to the above metrics, the manual test could deal with the answer whose semantic complexity and inherent ambiguity are relatively higher but its cost on time, money, and other resources is also higher.

Although many metrics have been proposed to evaluate the performance of visual question answering models on open-ended questions, both strengths and weaknesses existed in all metrics. The researcher should choose the metrics referring to the features of the dataset, the bias that existed in the dataset, and the expense that they could afford as well as pay more attention to design better metrics based on the current works.

IV MODELS

*A. Improve Efficiency*

*1) BLOCK Model[13]*

Bilinear model is an extension based on linear models, which enables two inputs, and it has become one of the most popular methods among those existing nowadays. However, when the input dimensions grow enormously, learning becomes difficult, due to the doubled number of parameters for input.

The idea of BLOCK is to decompose the original input into smaller block-terms that comprise the only fraction of original input, in order to reduce the number of parameters. Then, block-terms will be merged with a fusion parameterized by the block-superdiagonal tensor. Block-terms can be widely applied to the existing VQA model.

In practice, BLOCK model performances better in both time complexity and accuracy, compared to several preceding models, such as MCB [49], MFB [50] and MFH [51]. Moreover, BLOCK model also performances better in the fields of VRD.

*2) Grid Features Approach[14]*

The well-known bottom-up attention has surpassed vanilla grid-based convolutional features in the fields of VQA, but the involved contribution of regions remains inconclusive.

The grid features approach aims to convert region features to grids. Specifically, instead of using $14 \times 14$ RoIPool-ed features in Faster R-CNN model, $1 \times 1$ RoIPool is supposed, which decomposes three-dimensional tensors into single vectors. Moreover, in order to keep pre-trained convolutional layers still work for the change of inputs, only parts up to $C_5$ of ResNet[52] is kept.

In practice, Faster R-CNN with a ResNet-50 backbone pre-trained on ImageNet by default and co-attention model implemented in Pythia [53, 54] are used as setups. Compared to the model involving widely-used bottom-up region features, the adapted one with grid features results in higher accuracy with much fewer time consumptions.

*3) Differential Adaptive Computation Time [55]*

Differential Adaptive Computation Time (DACT) is an emerging attention-based algorithm that achieves the feature of end-to-end differentiable. To achieve such features, the requirement of the model is confined to those which can be decomposed into a series of submodels. Different from the general process, for each submodel, to prevent the subsequent models' influence of altering answers, sigmoidal outputs will be outputted besides the targeted outputs, and a probability will be computed based on the sigmoidal outputs. Then, each accumulated sub-output will be obtained by combining the previous submodel's output with the current submodel's output with "probability punishment". Recursively, the final output will be observed at the last submodel.

In practices, DACT is applied to CLEVR data set. Compared to MAC network, the model with DACT outperforms the model with MAC with a similar computation cost. However, such performance only holds within 12 iterations. Moreover, DACT iterates much fewer times when the question is relatively easy but more in the other case.

*B. Improve Prediction Accuracy*

*1) PLAC model [56]*

Most of the recent progresses on VQA are based on RNNs with attention. The successes gained are remarkable, but the longtime consumption is still a problem, and, due to the nature of RNN model, difficulties in modeling long-range dependencies remain unsolved.

The new idea of Positional Self-Attention with Co-Attention (PLAC) model contributes to improve computational efficiency and to derive long-rang dependencies. PLAC model is comprised of three key parts - Video-based Positional Self-Attention Block (VPSA), Question-based Positional Self-Attention Block (QPSA) and Video-Question Co-Attention Block (VQ-Co). Both VPSA and QPSA conduct video pre-processing and question pre-processing steps, respectively. Consequently, VPSA obtains frame features, and QPSA obtains both word level and character level features. Finally, VQ-Co is applied to boost the question answering performance.

In practice, PLAC model improves the performance of a high-level concept word detector to generate a list of concept words.

*2) Adversarial Learning in VQA[57]*

In the past few years, focuses on improving accuracies for VQA are within the model-level; in other words, reducing the biases from learning is the target. The absence of studying in data augmentation problem for VQA may impede the further development of VQA.

The recently studied adversarial learning for VQA problems aims to add the least amount of examples to inputs to achieve the desired misclassification. In order to avoid the effects of answers, only raw inputs (image and answer) are manipulated. With regard to manipulating images, the IFGSM[11] is used, which can produce harmful adversarial examples. Considered the risk of destroying grammar and semantics, the paraphrasing model [12] is applied to text inputs. Those adversarial examples are then treated as training samples and trained with a loss function to control the relative weight of adversarial examples.

In practice, such a method, applying on VQAv2 validation, test-dev, and test-std sets, outperforms BUTD vanilla training setting in validation, especially when the training set size tends to be small.

*3) TRRNet Model [58]*

TRRNet model is an attention-based model that follows the general VQA training process. The model is connected by TRR unit, which comprises four parts: root attention, root to leave attention passing, leaf attention, and message passing module.

The set of image features, bounding box features, and question features are firstly used to produce root attentions, which generates attention maps for object-level visual features based on languages and generates fused visual features. Then, outputs from root attention are processed at the root to leaf attention passing to generate pairwise relations, in which multi-head hard attention [59] are involved to select relevant objects. Then, the leaf attention is used to process object relation reasoning from the previous part, which outputs an attention map and a merged relation feature. At last, in the message passing module, the relation features from the previous step and object-level features from root attentions are fused. Readout layer will contribute to producing the final answer.

In practice, the model is applied to GQA [60], VQAv2, and CLEVR data set, with pre-trained Faster-RCNN, Bert word embedding and GRU. Compared to basic attention models, TRRNet model outperforms both when attention models are strong and weak, with significant improvement of Y/N questions' accuracy.

*C. Reducing Bias*

*1) RUBi Learning Strategy[61]*

While VQA makes great progress in recent years, the unimodal biases to provide the correct answer without using the image information are always a problem. Such biases may lead to a decrease in the performance of the model when evaluating data sets diverse from the trained one.

The RUBi learning strategy, inspired by the question-only model, contributes to reducing biases in VQA models. Different from the traditional progress of directly merging two inputs modalities, RUBi learning strategy takes only one of two modalities as input and adapts a question-only branch to capture the question biases. Then the outputs from the question-only branch will be merged with a mask to balance the scores of answers. Consequently, the loss will be lowered for biased examples.

In practice, architecture with Bilinear BLOCK fusion trained with RUBi learning strategy increases the average overall accuracy and decreases the standard deviations, compared to other relatively traditional models (e.g., UpDn[62], MuRel) on the data set VQA-CP v2.

*2) VCTREE Model [63]*

Despite the traditional RNN-based training model, the tree-based model also contributes to the fields of VQA task. The proposal of VCTREE has significant strengths in explaining parallel and hierarchical relationships among objects and permitting task-specific message passing among objects.

VCTREE model comprises four key steps. The visual features are firstly detected through Faster-RCNN. Then, a learnable matrix, approximate task-dependent validity between pairs of objects, is applied to build up VCTREE, which will be further employed by Bi-Tree LSTM to encode contextual cues. Finally, those encoded contexts will be decoded through the VQA model.

In practice, VCTREE performs better relative to other previously discovered methods (e.g., Count, MLB) on the data set VQA 2.0. However, when the data set involves biases, the error rate resulted from the model is still a problem to be solved in the future.

*3) Visually-Grounded Question Encoder [64]*

Visually-Grounded Question Encoder (VGQE) derives from adopted RNN based question encoding scheme in VQA, compromises two parts – Visually-Grounded Word (VGW) embedding part and traditional RNN cells. It pre-takes word embedding of questions, finds the corresponding relevant visual features of images, and generates a visually-grounded question word embedding vector, at the stage of VGW. Then, the outputted vector is exposed to RNN to further encode sequence information.

In practice, VGQE is tested against state-of-the-art bias reduction techniques in VQA-CPv2. Similar to the results from RUBi model, the overall accuracy, specifically among questions involving number-based answers.

*4) Counterfactual Samples Synthesizing Training Scheme[65]*

Counterfactual Samples Synthesizing (CSS) training scheme compromises three steps – training VQA models with original models, generating counterfactual samples, and training model VQA models again with counterfactual samples. To generate counterfactual samples, two steps are followed: (1) to use V-CSS or Q-CSS [65] to calculate contributions of each object features to ground-truth answers; (2) to use CO_SEL and DA_ASS [65] to combine counterfactual visual inputs and original questions to generate counterfactual samples.

In practice, the CSS training scheme is applied to VQA-CP data set with model UpDn[66], PoE [23], RUBi[61], and LMH. The result indicates the accuracies are improved significantly among the four models. Also, a similar and more significant result can be observed when the scheme is applied to VQA-CP v1 data set with model LMH.

## V. APPLICATION

### A. ROLL[16]

The aim of VQA application has not only limited to images, but also videos. Inspired by human behavior of constantly reasoning over the communications and actions through the storyline in the movie, model ROLL aims to leverage tasks of dialog comprehension, scene reasoning, and storyline recalling, with access to external resources to retrieve contextual information.

The task is achieved by 3 branches corresponding to read, observe and recall. In the reading branch, information is extracted from subtitles, and is then fed into the reading transformer to obtain read scores. In the observe branch, instead of directly applying pre-trained models and standard video captioning models, unsupervised video descriptions are used to create a video scene graph, which includes four nodes - character, place, object relation and action, and six edges; then, corresponding scene descriptions will be generated based on scene graphs. The resulting scene descriptions will later be used as inputs to feed into the observing transformer to compute observe scores. In the recall branch, inspired by KnowIT VQA [43], the video story it belongs to is firstly identified, and similarities between each frame in the scene and the whole frames are computed, and the video of most similar frame are kept; the identifier is then fed into recalling transformer to output the recall score for answers. Finally, the output from the three branches are used to output the predicted answers.

In practice, the ROLL model is applied to data set KnowIT VQA and TVQA+, using BERT uncased base model with pre-trained initialization for transformers. Compared to model ROCK[43], ROLL outperforms on the knowledge-based samples and containing more semantic information when comparing visual representations, while still underperforms when against human performance.

### B. Location Aware Graph Convolution Network

This paper[67] proposed a new representation for objects in the videos and their spatial and temporal relationships through a fully-connected graph to help the model answer the questions that require considering both temporal locations of objects and interactions between humans and objects. In the location-aware Graph Convolutional Networks, features are firstly detected by a multilayer perceptron and a full-connected graph where its node corresponds to an object and the relationships between objects are represented by the edges is created subsequently. Besides, nodes also contain information about spatial and temporal locations of objects. Through this method, the objects would transfer information through edges to interact with each other to obtain the region features when the graph convolution is implemented. The output would be combined with a representation of the question to generate the answer. The cross-modality representation consists of the image feature, the weighted question feature, and the element-wise product of the image feature and the weighted question feature. This model was tested and compared with the-state-of-art methods on the TGIF-QA, Youtube2Text-QA, and MSVD-QA dataset by mean square error and accuracy. On the TGIF-QA dataset, the performance of the L-GCN is better than ST-VQA, Co-Mem, PSAC and HME as well as its performance on the multiple-choice question is better than other models on Youtube2Text-QA dataset. It seems that this model obtained better performance on the video question answering task because it do not be affected by the irrelevant background content that the existing spatial temporal attention mechanism cannot avoid.

### C. ISVQA[9]

Traditional VQA aims to take pairs of single-image and questions as inputs to predict answers through models. The newly generalized task is to generalize traditional settings into multi-image settings; in other words, the inputs then become pairs of multi-images and answer, while the primary task remains the same. The predicted answers can be both open-ended and multiple-choice.

The model applied to such tasks derives from single image VQA models. It firstly processes each image separately and obtains features to predict the answer, based on the LSTM-attention module. Followed by the previous process, a distribution over the answers will be obtained. Then, images are stitched together and further processed by the adapted LXMERT model[9] to encode relationships between objects among images and predict the answer.

In practice, the accuracy remains to be improved with only a 40-50% accuracy rate. The method applies worse in video VQA models than that in image VQA models.

## VI. CONCLUSION

In conclusion, this paper reviewed the datasets, metric and models especially that works after 2018 and found that the research scope of visual question answering has been expanded from static and discrete images to the dynamic and continuous videos, even the 360-degree pictures and scientific diagrams were also explored by the researchers. Meanwhile, the attention of researchers has also been paid to the reasoning ability of models. Though achievements have been achieved in visual question answering tasks, several problems still exist. Firstly, language prior still has a negative influence on the visual question answering. In addition, statistical bias is difficult to reduce. Besides, for different types of questions, the existing metrics are not sufficient. Moreover, the multimodal fusion mechanism still needs refinement. Visual question answering is still a field worth exploring.